\documentclass[runningheads]{llncs}

\usepackage[T1]{fontenc}
\usepackage{graphicx}
\usepackage{amssymb,amsmath,array}
\usepackage{mathtools}
\usepackage{cleveref}
\usepackage{xcolor}
\usepackage{cite}
\usepackage{caption}
\usepackage{subcaption}
\usepackage{booktabs}

\newcommand{\etal}{\textit{et al}. }
\newcommand{\ie}{\textit{i}.\textit{e}.}

\begin{document}
\title{Improved Allergy Wheal Detection \\for the Skin Prick Automated Test Device}
\titlerunning{Improved Allergy Wheal Detection for the SPAT Device}

\author{
Rembert Daems\inst{1,2,3} \and
Sven Seys\inst{3} \and
Val\'erie Hox\inst{4} \and
Adam Chaker\inst{5} \and
Glynnis De Greve\inst{6} \and
Winde Lemmens\inst{7} \and
Anne-Lise Poirrier\inst{8} \and
Eline Beckers\inst{7} \and
Zuzana Diamant\inst{9} \and
Carmen Dierickx\inst{7} \and
Peter W. Hellings\inst{9,10} \and
Caroline Huart\inst{4} \and
Claudia Jerin\inst{5} \and
Mark Jorissen\inst{9,10} \and
Hanne Osc\'e\inst{6} \and
Karolien Roux\inst{10} \and
Mark Thompson\inst{11} \and
Sophie Tombu\inst{8} \and
Saartje Uyttebroek\inst{9} \and
Andrzej Zarowski\inst{6} \and
Senne Gorris\inst{3,12} \and
Laura Van Gerven\inst{9,10} \and
Dirk Loeckx\inst{3} \and
Thomas Demeester\inst{1}
}
\authorrunning{R. Daems \etal}

\institute{
Ghent University--imec, Ghent, Belgium \and
D2LAB, Ghent University, Ghent, Belgium \and
Hippo Dx, Aarschot, Belgium \and
Cliniques Universitaires Saint-Luc, Brussels, Belgium \and
Klinikum rechts der Isar, Munich, Germany \and
GZA Sint-Augustinus, Antwerp, Belgium \and
ZOL, Genk, Belgium \and
CHU Li\`ege, Li\`ege, Belgium \and
KU Leuven, Leuven, Belgium \and
UZ Leuven, Leuven, Belgium \and
Zurich University of Applied Sciences, Zurich, Switzerland \and
AZ Herentals, Herentals, Belgium
}

\maketitle

\begin{abstract}
\,

\textbf{Background:}
The skin prick test (SPT) is the gold standard for diagnosing sensitization to inhalant allergies.
The Skin Prick Automated Test (SPAT) device
was designed for increased consistency in test results, and captures 32 images to be jointly used for allergy wheal detection and delineation, which leads to a diagnosis.

\textbf{Materials and Methods:}
Using SPAT data from $868$ patients with suspected inhalant allergies, we designed an automated method to detect and delineate wheals on these images.
To this end, $10,416$ wheals were manually annotated by drawing detailed polygons along the edges.
The unique data-modality of the SPAT device, with $32$ images taken under distinct lighting conditions,
requires a custom-made approach.
Our proposed method consists of two parts: a neural network component that segments the wheals on the pixel level, followed by an algorithmic and interpretable approach for detecting and delineating the wheals.

\textbf{Results:}
We evaluate the performance of our method on a hold-out validation set of $217$ patients.
As a baseline we use a single conventionally lighted image per SPT as input to our method.

\textbf{Conclusion:}
Using the $32$ SPAT images under various lighting conditions offers a considerably higher accuracy than a single image in conventional, uniform light.

\end{abstract}

\section{Introduction}
\label{sec:introduction}
Inhalant allergies, such as pollen or house dust mite, are affecting one-third of the general population worldwide~\cite{sanchez2018importance}.
Local histamine releases are caused by the sensitization to specific allergens.
The skin prick test (SPT)~\cite{ebruster1959prick} is a diagnostic method where
a small amount of allergen is applied into the skin to provoke an allergic reaction, or wheal.
Together with in vitro diagnostic tests, the SPT is the gold standard for diagnosing allergic diseases~\cite{heinzerling2013skin,bernstein2008allergy}.

Recently, SPT was shown to be more sensitive than in vitro testing~\cite{gureczny2023allergy}.
However, the main drawbacks of SPT are the dependency of the accuracy on the expertise of the operator, which devices where used and the consistency of the execution of the test~\cite{mccann2002reproducibility,carr2005comparison}.

The \emph{Skin Prick Automated Test} (SPAT) device has been developed to bring standardization of the SPT procedure.
In brief, the patient is asked to hold the forearm against an armrest for a few seconds and 12 pricks will be applied simultaneously on the forearm.
After 15 minutes, the arm is repositioned in the SPAT device for capturing the images.
This is done at a position and orientation as close as possible to the original arm position during the prick procedure, using a visual cue.
The test is considered positive if the longest diameter of the skin reaction wheal exceeds $4.5 \text{mm}$~\cite{gorris2023reduced}.
The SPAT device offers less variability, more consistency of the test and less pain perceived by the patient~\cite{gorris2023reduced,seys2024skin,seys2023evaluation}.

\begin{figure}[!b]
    \centering
    \includegraphics[width=0.75\textwidth,trim=90 0 90 80,clip]{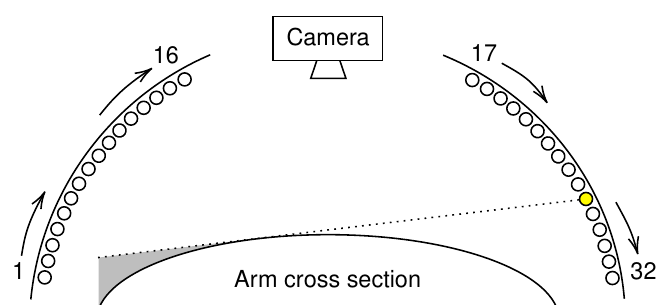}
    \caption{Cross view schematic of the SPAT camera and lighting system.
    Each of the $32$ images has its own illumination source, as indicated by $1$-$16$ on the left and $17$-$32$ on the right.
    As an example, $27$ is lit, and its corresponding shadow zone is shown in gray.
    The full-light image is captured by using the top six lights $14$-$19$, which produces uniform lighting on the arm.
    See~\Cref{fig:example-data-modality} for an example of the images.
    }
    \label{fig:spat}
\end{figure}
The SPAT device captures $35$ images for one SPT. These encompass 
$32$ images 
taken with distinct lighting conditions, where each image has its own, unique lighting source under a specific angle (\Cref{fig:spat}).
Three more images are control images,
\ie, one with no lighting, to check possible interference of ambient light sources outside of the SPAT,
and two full-light images where the light sources are configured to achieve uniform lighting.
One of the full-light images is used in this work as a baseline method, see \Cref{sec:results}.

\begin{figure}
    \centering
    \includegraphics[width=0.19\textwidth]{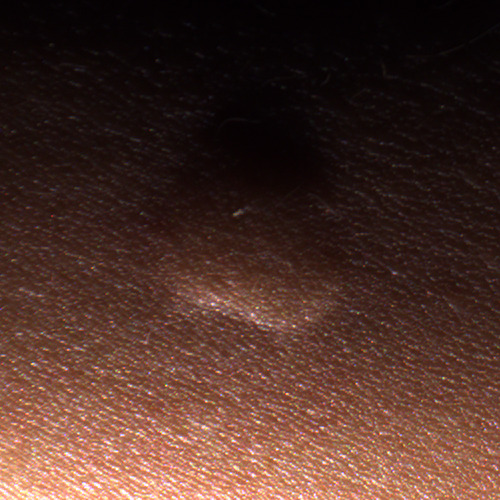}
    \includegraphics[width=0.19\textwidth]{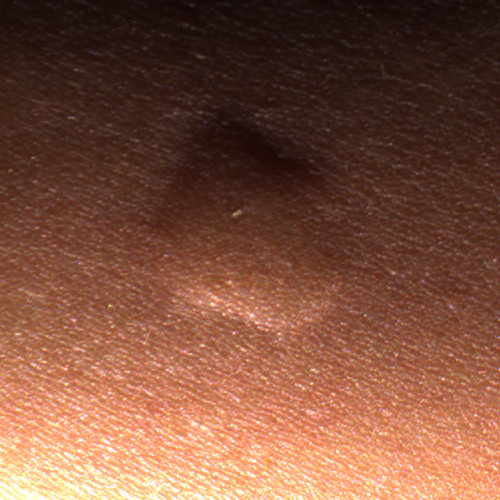}
    \includegraphics[width=0.19\textwidth]{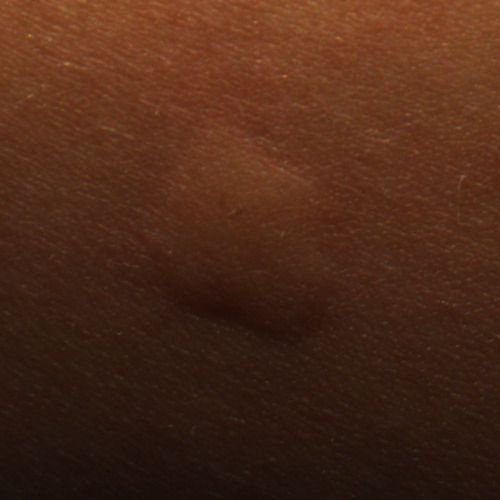}
    \includegraphics[width=0.19\textwidth]{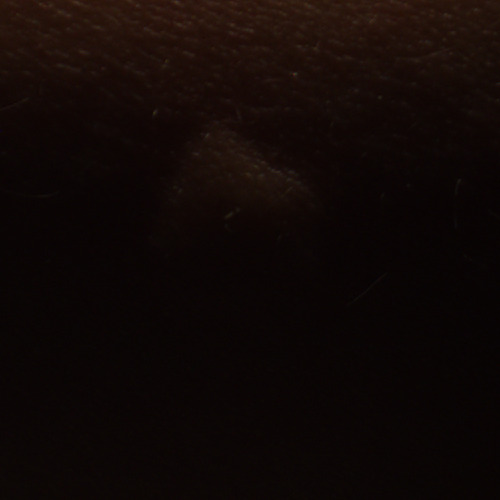}
    \unskip\ \vrule\ 
    \includegraphics[width=0.19\textwidth]{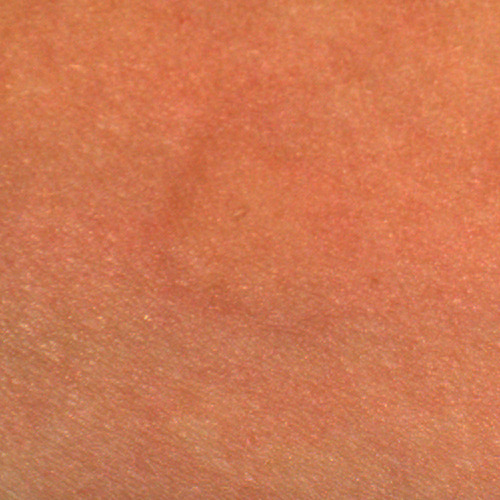}
    \caption{Zoomed-in view on one wheal to illustrate the specific data-modality of the SPAT.
    $32$ images are captured under distinct lighting conditions, here the same crop around a wheal of the first, third, 18th, 25th and a full-light image (from left to right) are shown as an example.
    It is obvious the first four images with the specific SPAT lighting contain more information about the wheal and its boundary than the last image with uniform lighting.
    Depending on the location of the wheal, a combination of several of these complementary images provides the necessary information to determine 
    the wheal shape.
    The full-light image is used for the baseline method (see \Cref{sec:results}).}
    \label{fig:example-data-modality}
\end{figure}

This data-modality is unique to the SPAT device and specifically designed to enhance the visibility of the wheals.
Distinct shadow lines at the edges of the wheals can be used to detect and delineate the wheals.
Depending on the location of a wheal, a combination of several of these complementary images provides the optimal information to determine the wheal shape.
\Cref{fig:example-data-modality} shows an example of the data-modality of the SPAT device.
In that sense, we want to leverage this unique data-modality to develop a method that can accurately and robustly detect wheals.
While other works have  developed methods to segment allergy wheals on conventional images~\cite{lee2024allergy,bulan2014improved,justo2016prick,szeliski2022computer},
we present an automated method to detect wheals that is compatible with the SPAT device.
It enables further standardization of the full chain of the SPT process, supporting physicians in the read-out of the test results.

\section{SPAT Data}
\label{sec:spat}
The data consists of $868$ adult patients with suspected inhalant allergies,
who underwent a skin prick test using a SPAT device between September 2023 and September 2024.
The patients were recruited in five Belgian hospitals and one German hospital.
All study participants provided written informed consent before inclusion in the study. 
Of these $868$ prick tests, $651$ ($75\%$) were used as training set for the deep learning segmentation model, and $217$ ($25\%$) were used as validation set to calculate the metrics presented in the Results section.
This train-validation split was done randomly, but stratified over the different hospitals, to ensure a balanced representation of the different hospitals in both sets.
On the in total $868$ prick tests, $10416$ wheals were manually annotated by drawing detailed polygons along the edges.

\section{Automated Detection of Wheals}
\label{sec:methods}
\begin{figure}[t]
    \centering
    \includegraphics[width=\textwidth]{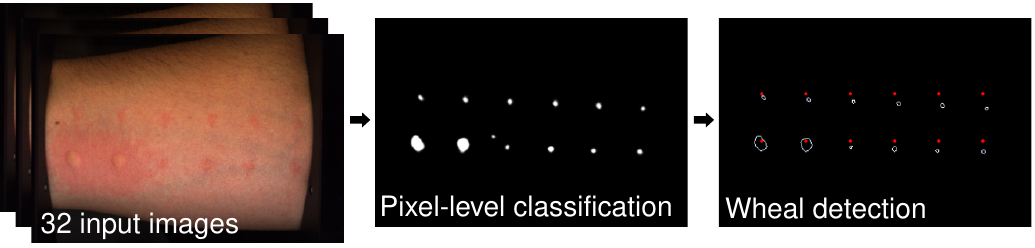}
    \caption{Schematic overview of our method. The $32$ input images are interpreted by the deep neural network 
    model that classifies the wheal regions on the pixel level. This leads to a segmentation map with the same resolution as the input images, and with values between $0$ and $1$ indicating wheal presence.
    In the second step, individual wheals are detected and paired to the known prick locations (red dots).
    Note that in this particular example, a small misclassified dot in the segmentation map is correctly discarded by the second step, because
    it could not be interpreted as a wheal.
    This shows the importance of the second step as a sanity check.
    Also note that the wheals are not exactly on the theoretical prick locations, because it is impossible to reposition the arm
    at the exact same location as it was during the pricks.
    This is taken into account when pairing the detected wheals to the known prick locations.
    }
    \label{fig:method-two-parts}
\end{figure}

Our method consists of two steps.
First, a deep neural network predicts whether individual pixels reflect an allergic reaction (\ie, belong to an allergy wheal).
Second, individual wheals are detected and paired with known prick locations.
See~\Cref{fig:method-two-parts} for a schematic overview.
\subsection{Step 1: pixel-level classification model}
\begin{figure*}[h]
    \includegraphics[width=\textwidth,trim=45 5 45 5,clip]{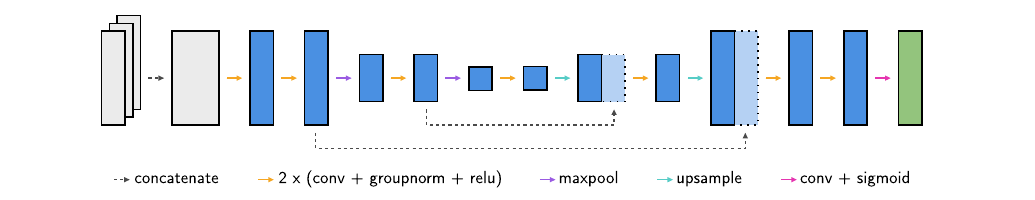}
    \caption{Architecture of the pixel-level classification model.
    Grey indicates the input images, blue indicates the hidden feature layers and green the output segmentation map.
    The height of the blocks indicate the spatial dimensions, which are halved after the maxpool operation (purple arrow) and doubled again after the upsample operation (cyan arrow).
    The dashed black arrows indicate concatenations: first of the $32$ input images along the channel dimension ($32 \times 3 = 96$),
    and later of the hidden feature layers that are concatenated at corresponding spatial resolutions while upsampling, inspired by the U-Net architecture~\cite{ronneberger2015u}.
    For the upsampling operation, we use bilinear interpolation.
    The yellow arrows indicate a double block of convolution layer, groupnorm~\cite{wu2018group} layer and ReLU activation.
    Finally, the pink arrow indicates a final convolution layer with $1$ output channel and a sigmoid activation to reach the required output of $[B, 1, H, W]$, inside the range of $(0, 1)$.
    All convolution layers are 2-d convolution operations with kernel size $3 \times 3$ and stride $1$.
    All the hidden layers in the model have $64$ features.}
    \label{fig:segmentation-architecture}
\end{figure*}

One of the main novelties in our method lies in the handling of the unique data-modality generated by the SPAT device.
One skin prick test generates $32$ images that are captured with distinct lighting conditions (see~\Cref{sec:introduction}).
This means we can not simply take an off-the-shelve or pre-trained deep learning segmentation model.
Specifically, the input to the segmentation model is a tensor with shape $[B, F, C, H, W]$, where $B$ is the batch size,
$F$ is the number of images for one prick test ($32$), $C$ is the number of channels ($3$ in this case, for color images), and $H$ and $W$ are the height and width of the image.
Our solution to handle this specific data-modality is simple yet effective.
We concatenate the $F$ dimension on the channel dimensions such that: $[B, F, C, H, W] \to [B, F C, H, W]$.
This renders a tensor with $32 \times 3 = 96$ channels, where the $32$ different images are simply concatenated along the channel dimension.
Because of this concatenation, our method is now compatible with standard deep learning layers.
The model further consists of 2-d convolution layers, group normalization layers~\cite{wu2018group} and ReLU activations,
structured as a U-Net to allow precise localization~\cite{ronneberger2015u}.
The output of this model is the same height and width as the input images, with one output channel.
This channel contains the predicted segmentation mask of the wheals, \ie, a value between $0$ and $1$ for every pixel in the image indicating the presence of a wheal.
See~\Cref{fig:segmentation-architecture} for the full neural network architecture.

The segmentation model is trained using binary cross-entropy loss.
Images are resized to $768 \times 512$ pixels, and the pixel values are normalized to the range $(0, 1)$.
We use the Adam optimizer~\cite{kingma2014adam} with a learning rate of $3 \times 10^{-4}$.
Our implementation uses PyTorch~\cite{paszke2019pytorch} and PyTorch Lightning~\cite{Falcon_PyTorch_Lightning_2019}.
We train the model on a single NVIDIA GeForce RTX 2080 Ti GPU, with a batch size of $4$.
Training is done for $64$ epochs, which takes approximately $34$ minutes.

\subsection{Step 2: algorithmic detection of wheals}
The second step in our method is to aggregate the pixel-level predictions obtained after the first step into measurable allergy wheals associated to the prick locations.
We specifically choose not to train a neural network end-to-end for detecting wheals directly from the pixel images.
Such a black-box approach would render interpreting the model's predictions very difficult. Instead, we opt for splitting up the pure pattern recognition part \emph{(step 1)} and the detection and delineation of wheals based on the pixel-level predictions \emph{(step 2)}. This allows for interpreting the model's decisions, in terms of pixel-level predictions (by inspecting step 1 outcomes), as well as their aggregation into consistent wheals (based on the white-box approach in step 2, as described below).

The pixel-level classification from step 1 is 
processed by first thresholding to a binary mask, and then applying a connected components algorithm to obtain distinct wheal regions~\cite{fiorio1996two}.
This renders a list of wheal candidates, where every wheal is represented by a list of pixel coordinates.
This list of wheal candidates is matched with the list of known prick locations.
Since we only allow one match per prick location, the allowed options are (1) pairing a prick location with a detected wheal, (2) not finding any detected wheal for a prick location, or (3) not finding any prick location to be associated with a detected wheal.

Since the patient's arm is repositioned in the SPAT device between the prick procedure and the image capturing (see~\Cref{sec:introduction}),
there can be some difference in the arms position and orientation.  This is taken into account by matching the predicted wheals to the theoretical prick locations using a parametric transformation.
We use a rigid transformation in the image plane, \ie, using three parameters: translation in $x$ and $y$ direction, and rotation.
We search for the optimal transformation that minimizes the distances between the predicted wheals and the known prick locations.
This is done separately for each prick test, with a 
grid search over these three parameters, as we found this the most robust method in practice.
After this global optimization, a greedy search results in pairs of matched predicted wheals and known prick locations.

\section{Results}
\label{sec:results}
For comparison, a baseline model is trained on one full-light image as input.
The neural network architecture is adapted so that the first layer accepts $3$ channels as input, instead of  $32 \times 3 = 96$ (see \Cref{fig:segmentation-architecture}).
All else in step 1 and step 2 staying equal allows us to ablate the SPAT specific $32$ image data-modality.
\subsection{Performance of the pixel-level classification model (step 1)}
We first present results on the first step of our method, the neural network pixel-level classification model.
We evaluate the performance using the Dice coefficient on the pixel level, which is a common metric for image segmentation tasks.
First, the output of the model is thresholded at $0.5$ to obtain a binary mask.
Then, the Dice coefficient is calculated between the binary mask and the ground truth annotation: $2 \text{TP} / (2 \text{TP} + \text{FP} + \text{FN})$,
where TP, FP and FN are the number of true positive pixels, false positive pixels and false negative pixels, respectively.
See~\Cref{tab:results} for results.

\subsection{Performance of the entire pipeline (step 1 and step 2)}
Additionally, we present results after the second step of our method, reflecting the final performance of the complete pipeline.
As explained in~\Cref{sec:methods}, the final output of our model is a list of detected wheals, paired to the known prick locations.
We evaluate the performance of this final output using the Intersection over Union (IoU) metric 
which is a common metric for object detection tasks~\cite{szeliski2022computer}.
The IoU is calculated as the area of overlap between the predicted wheal and the ground truth wheal, divided by the area of their union:
\begin{equation}
    \text{IoU} = \frac{\text{Area of intersection}}{\text{Area of union}} \, .
\end{equation}
Thus it is a normalized metric bounded by $0$ (the prediction has no overlap with the ground truth at all)
and $1$ (the prediction and ground truth match perfectly).
When a prick location is not matched to any detected wheal, we set the IoU to $0$.
For this analysis, we do not take into account the very small non-elevated prick marks that are always present, but clinically irrelevant.
We set a threshold at $15.9 \text{mm}^2$ for the ground truth area of the wheals to discern these small marks.
Under the assumption of a perfect circle, this area threshold is equivalent to a diameter of $4.5 \text{mm}$, which was shown to be a reasonable
threshold for clinical relevance~\cite{gorris2023reduced}.
Using this criterion, $474$ of the in total $2604$ wheals in the validation set are used for the IoU analysis.

\begin{figure}
    \centering
    \includegraphics[width=.384\textwidth]{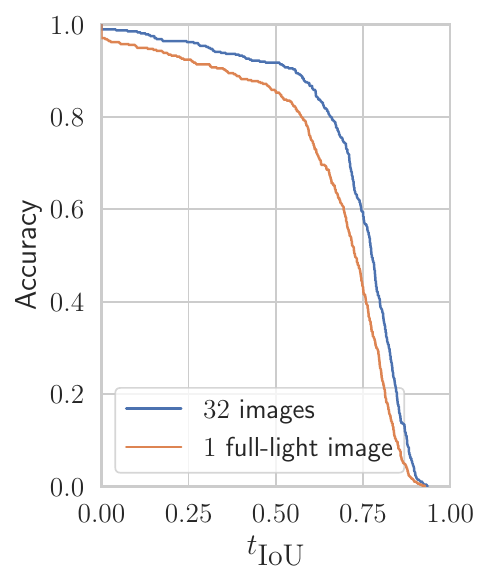}
    \includegraphics[width=.6\textwidth]{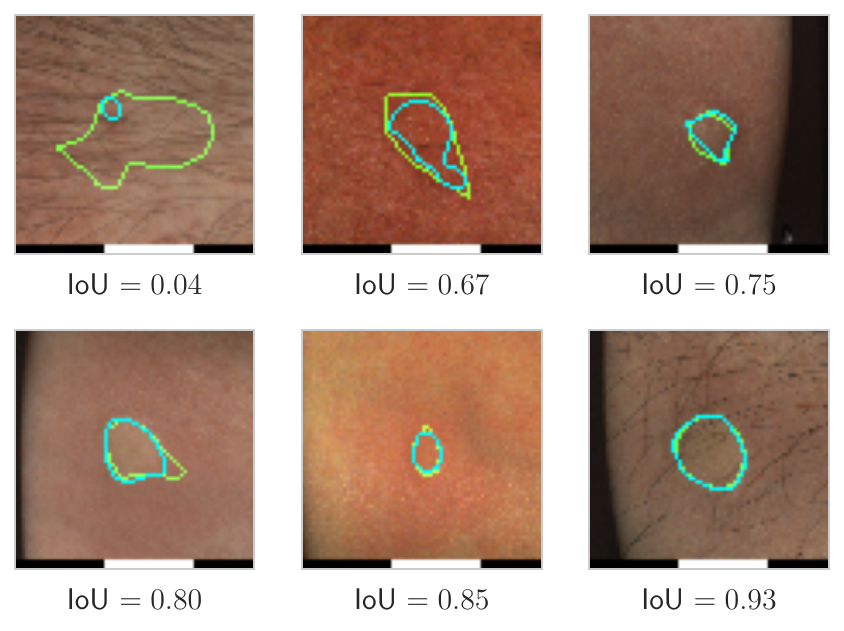}
    \caption{On the left we show the accuracy of the full pipeline on the validation set over the full range of IoU thresholds $t_\text{IoU}$.
    The IoU is $1$ if the detected wheal matches perfectly with the ground truth wheal, and $0$ if there is no overlap at all.
    Given a threshold $t_\text{IoU}$, a wheal detection is counted as accurate when its IoU is higher than the threshold.
    Over the full range our method is better than the baseline.
    On the right we show qualitative results of the wheal detection on six samples from the validation set.
    The wheals are ordered by the IoU metric, left to right, top to bottom.
    The samples are taken on equally spaced intervals of the IoU metric, ranging from the worst to the best result.
    Selecting the samples in this deterministic way shows the spread of the validation set
    and does not allow any cherry-picking.
    The green contour indicates the ground truth wheal, the cyan contour indicates the detected wheal.
    The scale is indicated on the bottom of every image.
    One black (or one white) block indicates $10 \text{mm}$.
    }
    \label{fig:iou}
\end{figure}

To present a clear metric, we define the IoU threshold $t_\text{IoU}$.
Given a choice of $t_\text{IoU}$, we define the accuracy as the ratio of wheals in the validation set that were detected with an IoU higher than $t_\text{IoU}$.
We present results over the full range of $t_\text{IoU}$ in~\Cref{fig:iou}, and on selected values in~\Cref{tab:results}.
\begin{table}
    \centering
    \caption{The Dice coefficient measures the performance of the pixel-level classification (step 1).
    The accuracy at varying IoU thresholds $t_\text{IoU}$ shows the performance of the full pipeline.
    Our method using the $32$ input images achieves considerably better results compared to the baseline using one full-light image.}\label{tab:results}
    \begin{tabular}{lc@{\hskip 1em}c@{\hskip 1em}c@{\hskip 1em}c@{\hskip 1em}c@{\hskip 1em}c}
    \toprule
    & Dice coefficient & \multicolumn{5}{c}{Accuracy (\%) at threshold $t_\text{IoU}$} \\
    \cmidrule(lr){3-7}
    & & $0.5$ & $0.6$ & $0.7$ & $0.8$ & $0.9$ \\
    \midrule

$32$ images & \textbf{0.787} & \textbf{91.8} & \textbf{86.7} & \textbf{74.3} & \textbf{38.8} & \textbf{2.3} \\
$1$ full-light image & 0.717 & 85.2 & 75.3 & 58.2 & 25.5 & 1.1 \\

    \bottomrule
    \end{tabular}
\end{table}

\subsection{Analysis of individual image contributions}
\begin{figure}
    \centering
    \includegraphics[width=\textwidth]{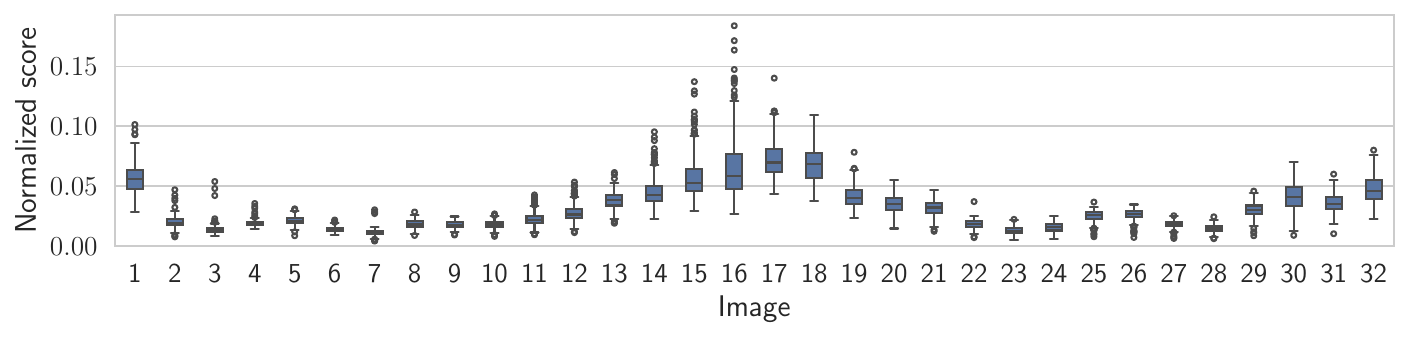}
    \caption{Distribution over the validation set of the normalized scores per input image.
    The score represents the sensitivity of a specific input image towards the output of the deep learning model.
    This indicates the relative importance of each of the $32$ lighting conditions for the pixel-wise classification of wheal regions.
    See~\Cref{fig:spat} for the numbering of the images.
    }
    \label{fig:score}
\end{figure}
Every pixel in each of the $32$ images has a certain sensitivity towards the result of the model~\cite{simonyan2014deep}.
This enables an analysis of the relative contributions, by calculating the gradient of the loss function with respect to the pixels of the input images.
We sum the squares of the gradients of all pixels per input image, which gives us $32$ score values per prick test.
\Cref{fig:score} shows the distributions of the normalized scores over the validation set.
This gives an indication of the relative importance of each of the $32$ lighting conditions. We observe the largest contributions from the light conditions illuminating the arm from the top center, as well as from the lights towards the bottom left and right (referring to the lights as indicated in~\Cref{fig:spat}).

\section{Conclusion and Future Work}
We present a method to automatically detect and delineate wheals on skin prick test images captured by the Skin Prick Automated Test (SPAT) device.
Our method consists of two parts: a deep learning model that handles the specific SPAT data-modality and classifies wheal regions on the pixel level,
and a white box algorithmic detection of the wheals and their boundaries.
This explicit two-step approach offers a high degree of interpretability of the model, which is important in a medical setting.
We evaluate the performance of our method on a hold-out validation set of $217$ patients and demonstrate the increased accuracy in allergy wheal detection based on the 32 images with varying lighting conditions from the SPAT, as compared to conventional single-image based prediction.

The data used in this study was collected as part of a broader clinical study (currently under analysis).
The machine learning component introduced in this paper
constitutes a vital component of the complete pipeline for allergy diagnosis using the SPAT,
assisting the physician in the read-out of the test.

\begin{credits}

\subsubsection{\discintname}
MJT received consulting fees for statistical advice for the study.
RD, SFS and DL are employees of Hippocreates BV.
RD, SFS, DL, SG and LVG hold shares of Hippocreates BV.
\end{credits}

\bibliographystyle{splncs04}

\begin{footnotesize}
\bibliography{references}
\end{footnotesize}

\end{document}